\documentclass[runningheads]{llncs}

\usepackage{enumitem}
\setlist{nosep, leftmargin=14pt}

\usepackage{graphicx}
\usepackage{float}
\usepackage[english]{babel}
\usepackage[utf8]{inputenc}
\usepackage{array, multirow, graphicx}
\usepackage{amsmath, amsfonts, xcolor}
\usepackage{etoolbox}
\usepackage{nicematrix}
\usepackage{tikz}
\usepackage{xspace}
\usepackage{bm}
\usepackage{unicode}
\usepackage[flushleft]{threeparttable}
\usepackage{makecell}
\usepackage{enumitem}
\usepackage{mwe}
\usepackage[fontsize=10pt]{fontsize}
\usepackage{hyperref}
\usepackage{academicons}

\newcommand{\etal}{et al.}
\newcommand{\orcid}[1]{\href{https://orcid.org/#1}{\textcolor[HTML]{A6CE39}{\aiOrcid}}}

\usepackage{bm}
\usepackage{amssymb}

\usepackage{algorithmic}
\usepackage{algorithm}
\usepackage{url}

\usepackage{multirow}

\def\T{{\mathrm{T}}}

\def\T{{\mathrm{T}}}

\def\s{{\mathbf s}}
\def\t{{\mathbf t}}
\def\S{{\boldsymbol{\xi}}}
\def\T{{\boldsymbol{\varphi}}}
\def\temp{ T}
\DeclareMathOperator*{\argmax}{arg\,max}

\def\bx{{\mathbf x}}
\def\by{{\mathbf y}}

\def\bX{{\mathbf X}}

\begin{document}
\title{Vision Transformers for Small Histological Datasets Learned through  Knowledge Distillation}

\author{Neel Kanwal \inst{1}*\href{https://orcid.org/0000-0002-8115-0558}{\includegraphics[scale=0.01]{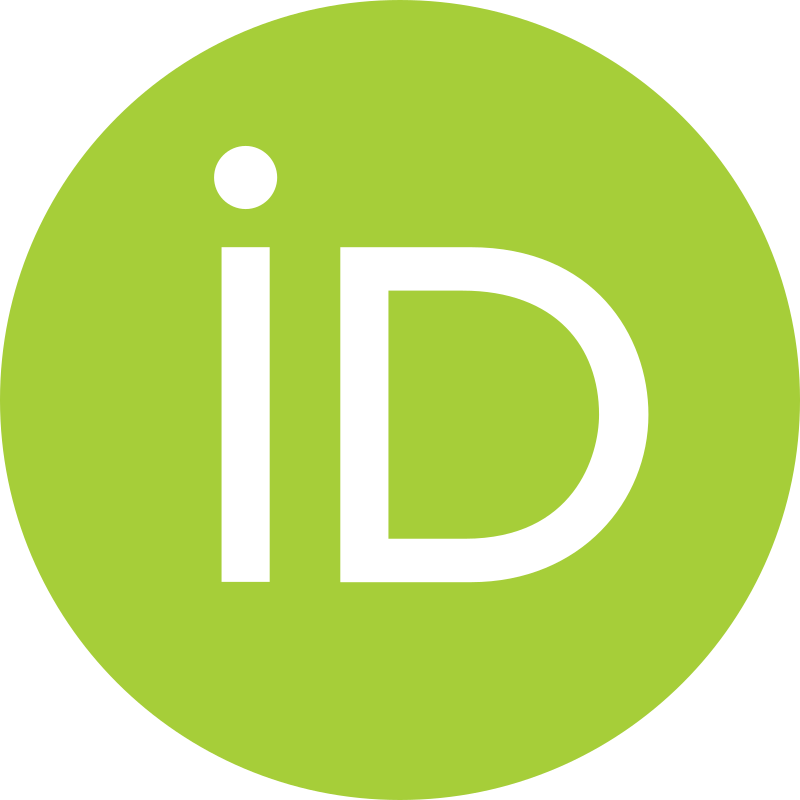}} \and 
    Trygve Eftestøl\inst{1} \and 
    Farbod Khoraminia\inst{2} \and
    Tahlita CM Zuiverloon\inst{2} \and
    Kjersti Engan\inst{1}\href{https://orcid.org/0000-0002-8970-0067}{\includegraphics[scale=0.01]{images/orcid.png}} }

\authorrunning{Kanwal et al., Vision Transformers for Artifact Detection} 
\titlerunning{Kanwal et al., Vision Transformers for Artifact Detection}

\institute{Department of Electrical Engineering and Computer Science, University of Stavanger, Norway 
\and Department of Urology, University Medical Center Rotterdam, Erasmus MC Cancer Institute, Rotterdam, The Netherlands \\
*Corresponding author: neel.kanwal@uis.no
}

\maketitle  
\vspace{-2em}
\begin{abstract}
Computational Pathology (CPATH) systems have the potential to automate diagnostic tasks. However, the artifacts on the digitized histological glass slides, known as Whole Slide Images (WSIs), may hamper the overall performance of CPATH systems. Deep Learning (DL) models such as Vision Transformers (ViTs) may detect and exclude artifacts before running the diagnostic algorithm. A simple way to develop robust and generalized ViTs is to train them on massive datasets. Unfortunately, acquiring large medical datasets is expensive and inconvenient, prompting the need for a generalized artifact detection method for WSIs.
In this paper, we present a student-teacher recipe to improve the classification performance of ViT for the air bubbles detection task. ViT, trained under the student-teacher framework, boosts its performance by distilling existing knowledge from the high-capacity teacher model. Our best-performing ViT yields 0.961 and 0.911 F1-score and MCC, respectively, observing a 7\% gain in MCC against stand-alone training. The proposed method presents a new perspective of leveraging knowledge distillation over transfer learning to encourage the use of customized transformers for efficient preprocessing pipelines in the CPATH systems. 
 \vspace{-2em}

\end{abstract}
\vspace{-2.75em}

\section{Introduction} \label{sec:introduction}
\vspace{-0.25em}
\vspace{-0.25em}
Histological examination of tissue samples is conducted by studying thin slices from a tumor specimen mounted on a glass slide. 
During the laboratory procedures, the preparation of glass slides may introduce artifacts and variations causing loss of visual ~\cite{kanwal2022quantifying,taqi2018review}. Artifacts, such as air bubbles, occur when air is trapped under the cover slip due to improper mounting procedure~\cite{kanwal2022devil}. Eventually, the presence of air bubbles leaves an altered and fainted appearance~\cite{kanwal2022devil,taqi2018review}. During the manual assessment, pathologists usually ignore regions containing artifacts as they are irrelevant for diagnosis.

Computational Pathology (CPATH) systems are automated systems working with a digitized glass slide, called Whole Slide Image (WSI), as input. CPATH systems have the potential to automate diagnostic tasks and provide a second opinion or localize the Regions of Interest (ROIs)~\cite{kanwal2023detection}.  
Different types of artifacts, like air bubbles, might be present on the WSI~\cite{kanwal2022devil} and can deteriorate diagnostic CPATH results if included in the analysis. Therefore it has been proposed to detect and exclude artifacts as a first step before using more relevant tissue in a diagnostic or prognostic system~\cite{kanwal2022quantifying,kanwal2022devil}.  The detection and exclusion of artifacts can be regarded as (a part of) a \emph{preprocessing pipeline}, which also might include color normalization and patching~\cite{kanwal2022devil}.  A complete preprocessing pipeline should detect folded tissue, damaged tissue, blood, and blurred (out of focus) areas, as well as air bubbles~\cite{kanwal2022devil}.  This might be done by an ensemble of models, one for each artifact, or by a multiclass model.   In this paper, we consider detecting \emph{air bubbles} artifact, which is not given much attention in the literature. 

Deep Learning (DL) methods have shown promising results in various medical image analysis tasks~\cite{fuster2022invasive,tomasetti2022multi}, and can be used for detecting artifacts in a preprocessing pipeline. 
Supervised learning for generalized DL models requires a significant amount of data and labels. 
In CPATH literature, little effort has been made to annotate artifacts; thus, 
publicly available datasets for histological artifacts are unavailable.
Transfer Learning (TL) has been widely used for medical images to deal with the lack of labeled training data~\cite{golatkar2018classification,noorbakhsh2020deep}. TL methods use the existing knowledge, such as ImageNet~\cite{imagenet} weights, and fine-tune the model for a different task. Although TL on ImageNet weights is useful to cope with a lack of data, ImageNet weights are mostly available for complicated Deep Convolutional Neural Networks (DCNN) architectures and carry a strong texture bias~\cite{geirhos2018imagenet}.  However, such DCNNs are typically computationally complex, whereas a preprocessing pipeline, being a first step prior to diagnostic or prognostic models, should have generalized and  efficient DL models with high throughput.  This is especially true  with an ensemble of DCNN models for the different artifacts.

After the success in natural language processing tasks, \emph{transformers} have been given attention for vision tasks~\cite{dosovitskiy2020image,kanwal2022attention}. Vision Transformers (ViTs), using a convolution-free approach, have surpassed DCNNs in accuracy and efficiency on image classification benchmarks~\cite{bhojanapalli2021understanding,dosovitskiy2020image}. Unlike the convolution layer in DCNNs, which applies the same filter weights to all inputs, the multi-head attention~\cite{vaswani2017attention} in ViTs attends to image-wide structural information~\cite{naseer2021intriguing}. Interestingly, ViTs are also shown to be more robust and generalized than DCNNs~\cite{bhojanapalli2021understanding,naseer2021intriguing}; Unfortunately, the robustness and generalizability come from training on extremely large datasets~\cite{bhojanapalli2021understanding,dosovitskiy2020image,Deit}, which contrasts with the biomedical scenario.
These limitations bring us to the question: \emph{how can we train generalized ViTs on a small histopathological dataset?}. 

\begin{figure}[ht!]
    \centering
\includegraphics[width=1\textwidth]{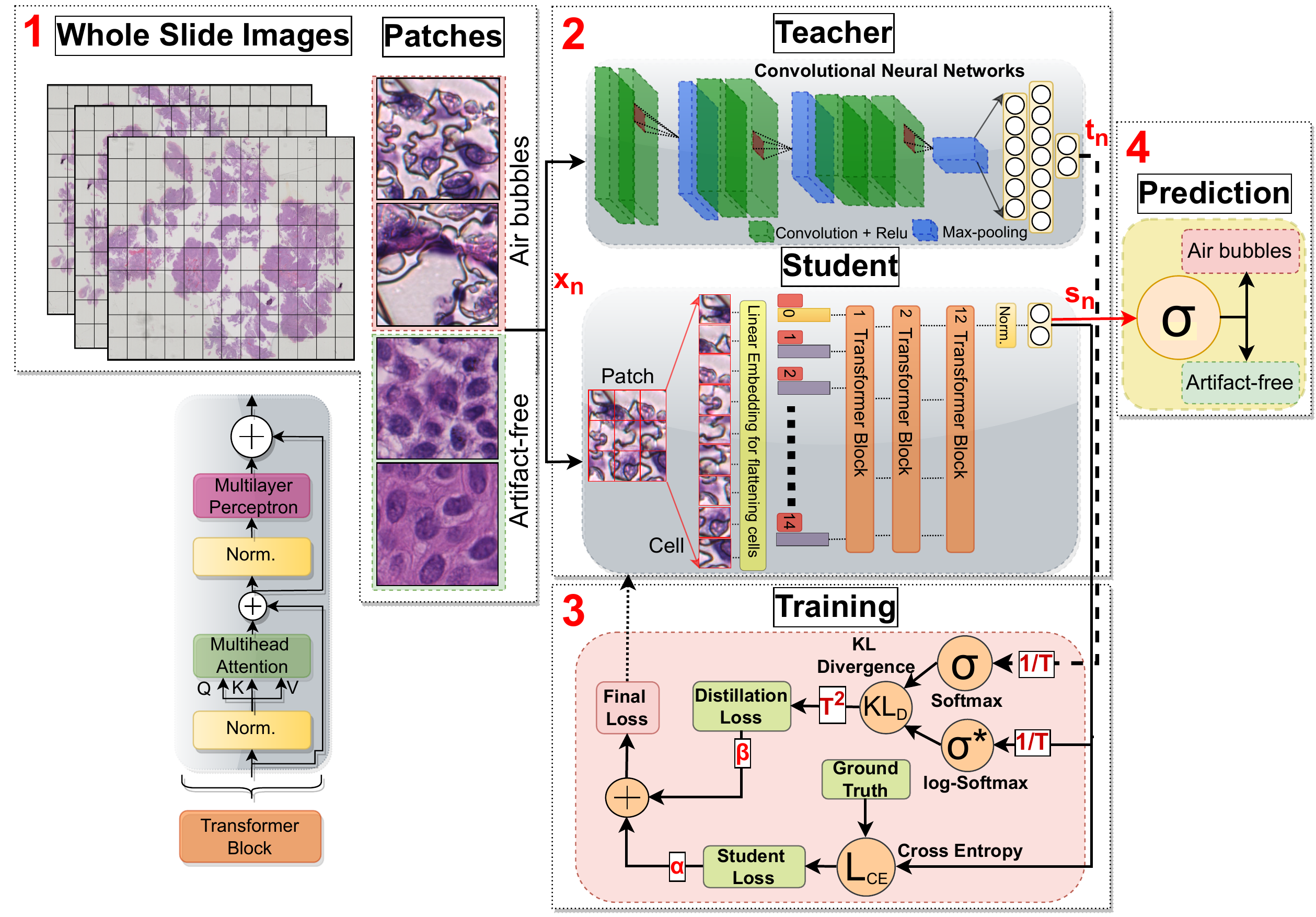}
    \vspace{-1.5em}
    \caption{\textbf{An overview of our proposed air bubbles detection  method by knowledge distillation:} Predefined size patches for air bubbles and artifact-free classes are extracted from the WSI. A ViT student model is trained with the help of a DCNN teacher model by leveraging the transference of knowledge during the training process. The student-teacher recipe weights the teacher and student's outputs by the temperature ($\temp$). The overall training objective is to minimize the final loss, which is a linear combination of student loss and distillation loss. Finally, the student model is used to perform predictions for binary air bubbles detection task. \vspace{-1em}}
    \label{fig:intro}
\end{figure}

One possible answer lies in Knowledge Distillation (KD)~\cite{hinton2015distilling}, which transfers knowledge from a usually large teacher model to another, typically smaller, student model.  Motivated by the KD idea, we present a student-teacher recipe, as shown in Fig~\ref{fig:intro}. 
We propose to use KD in combination with TL for detecting air bubbles on WSIs using a small training set. In short, we let the teacher model be a complex ImageNet  pretrained DCNN, and using KD, we train a small student model, which is a ViT. In the inference stage, we only need the small ViT, which is computationally efficient enough for a preprocessing pipeline implementation. 


Our contributions in this paper can be summarized as follows:

\begin{itemize}[wide]
\item We train several state-of-the-art DCNNs and ViTs to compare their performance on a binary air bubbles detection task. Later, we choose suitable architectures to test our student-teacher framework. 
\item We conduct an in-depth comparison by initializing models with and without ImageNet weights and training ViT under a standalone vs. a student-teacher framework. 
We also assess the improvements in ViT`s generalization capability over ImageNet transfer learning. 
\item We run extensive experiments to test the student ViT's performance under different teacher models and distillation configurations on unseen data.  
\end{itemize}
\vspace{-0.25em}

\vspace{-0.5em}
\section{Related Work}
\vspace{-0.5em}\label{sec:relatedwok}

{\bf Artifact and air bubbles detection}: 
The detection of histopathological artifacts has largely been overlooked during the development of CPATH systems, and the literature on air bubbles is scarce.  
Shakhawat \etal~\cite{Hossain2018}, in their quality evaluation method, detected air bubbles in two steps. First, the non-overlapping affected patches were detected using a Support Vector Machine (SVM) classifier. Later, the remaining patches with fainted appearance were separated using handcrafted Gray-level Co-occurrence Matrix (GLCM) features. This work was later extended in~\cite{Shakhawat2020}, where a pretrained VGG16~\cite{vgg16} network was used to compare the handcrafted features against the CNN-based method. Their experiments concluded that handcrafted features provide stable classification, but their evaluation was based on a relatively smaller dataset. Recently, Raipuria \etal~\cite{raipuria2022stress} performed stress testing for common histological artifacts, including air bubbles, using a vision transformer~\cite{Deit} and a ResNet~\cite{resnet} model. Though, MobiletNet~\cite{mobilenet} and VGG16~\cite{vgg16} have been popular DCNN choices for artifact detection~\cite{kanwal2022quantifying}. DCNNs are found to be less robust than ViTs and exhibit strong texture bias~\cite{naseer2021intriguing,raipuria2022stress}.

\vspace{0.25em}
{\bf Knowledge Distillation (KD)}: 
Originally proposed by Hinton \etal~\cite{hinton2015distilling} for model compression, KD sought to extract knowledge from an ensemble of CNN experts to a smaller two-layer CNN generalist network to make it perform equally well. In short, KD aims to train a small student model under the guidance of a complicated teacher model, where the student model optimizes its learning by absorbing the hidden knowledge from the teacher. This transference of knowledge can be accomplished by minimizing output logits of student and teacher networks through some distillation methods, such as logit-based, feature-based, and relationship-based distillation methods~\cite{meng2021knowledge}. 

KD helps make computationally friendly deployment algorithms, making it interesting for many biomedical imaging algorithms. Lingmei \etal~\cite{lingmei2021noninvasive} proposed a CNN model for glioma classification. They used the KD approach to compress the model and make it suitable for deployment on medical equipment. Salehi \etal~\cite{salehi2021multiresolution} used a VGG16~\cite{vgg16} cloner network to calculate multi-level loss from a source network for detecting anomalies. Their method relied on distilling intermediate knowledge from the ImageNet pretrained source network. In a similar approach, He \etal~\cite{he2021self} used the KD technique to boost the performance of CNN for ocular disease classification. They used fundus images and clinical information to train a ResNet~\cite{resnet} teacher  first and used only the fundus images to train a similar student network later. Guan \etal~\cite{guan2021mri} detected Alzheimer's disease by leveraging multi-modal data to train a teacher network. Their distillation scheme improved the prediction performance of the ResNet~\cite{resnet} student using a single imaging modality. 

However, all these works focused on using only CNN as a student network and did not explore the effects of different configurations and teacher networks on the final classification outcome. In addition, the use of KD for histological artifacts has not been investigated yet.  


\vspace{-0.25em}

\vspace{-1.5em}
\section{Data Materials and Method} \label{sec:method}
\vspace{-1.5em}
Fig.~\ref{fig:intro} provides an overview of our air bubbles detection method using KD~\cite{hinton2015distilling} in a student-teacher recipe. We exploit KD for data-efficient training by leveraging the transference of knowledge from the teacher model to the student model. Our proposed method uses a complex DCNN as the pre-trained teacher and a small ViT as the student when a small histological dataset is available.  We are doing a logit-based distillation~\cite{meng2021knowledge} since our teacher and student models are very different. The steps of our method are further described below. 

\vspace{-0.5em}
\subsection{Dataset}
\vspace{-0.25em}
The air bubbles dataset was prepared from 55 bladder biopsy WSIs, provided by Erasmus Medical Center (EMC), Rotterdam, The Netherlands. The glass slides were stained with Hematoxylin and Eosin (H\&E) dyes and scanned with Hamamatsu Nanozoomer at $40\times$ magnification. WSIs are stored in \emph{ndpi} format with a pixel size of 0.227~{$\mu$}m $\times$ 0.227~{$\mu$}m. These WSIs were manually annotated for air bubbles and artifact-free tissue by a non-pathologist who has received training for the task. To prevent data leakage, the dataset was later split into 35/10/10 training, validation, and test WSIs, respectively. 

\vspace{-0.75em}
\subsection{Foreground Segmentation and Patching}
\vspace{-0.5em}
Let $I_{\text{WSI}(i)}^{40x}$ correspond to a WSI at magnification level 40x (sometimes referred to as 400x).  $I_{\text{WSI}}^{40x}$ are very large gigapixel images, and it is not feasible to process the entire WSI at once. As such, all  CPATH systems resort to patching or tiling of the image, or the ROI in the image, before further processing. Let ${\cal T} : I_{\text{WSI(i)}\in R}^{40x} \rightarrow \{{\bf x}_{j}^{i}; j=1\cdots J \}$ represent the process of patching a ROI denoted by $R$ of the image $I_{\text{WSI}(i)}^{40x}$ into a set of $J$ patches, where  $\bx_j^{i}\in\mathbb{R}^{W\times H \times C}$ and $W$, $H$, $C$ present the width, height, and channels of the image, respectively. In the patching process, foreground-background segmentation was performed first by transforming (Red, Green, Blue) RGB images to (Hue, Saturation, Value) HSV color space. Later, Otsu thresholding was applied to the value channel to obtain the foreground with tissue. The extracted foreground was later divided over a non-overlapping square grid, and patches with at least 70\% overlap to the annotation region ($R$) were extracted.   

Let $\mathcal{D}=(\bX, \by) = \{(\bx_n,\by_n)\}_{n=1}^{N}$ denote our prepared dataset of N patches from a set of WSIs and $y_n\in\{0,1\}$ is the binary ground truth for the $n$-th instance, where 1 indicates a patch within a region marked as air bubbles. Fig~\ref{fig:intro} (step 1) shows the patches $\bx_{n}$ of $224\times224\times3$ pixels with air bubbles and artifact-free classes obtained from a WSI at 40x magnification.

\vspace{-0.75em}
\subsection{Selecting Student-Teacher Architectures}
\vspace{-0.5em}
Let`s symbolize the student model $\S$ with parameters $\theta$ providing the prediction output logits $\s_{n}=\S_{\theta}(\bx_{n})$, and correspondingly,  the teacher model $\T$ parameterized by $\phi$ providing the output logits $\t_{n}=\T_{\phi}(\bx_{n})$.

Our student model is a ViT, similar to the pioneering work~\cite{dosovitskiy2020image}, which leverages multi-head self-attention mechanism~\cite{vaswani2017attention} to capture content-dependant relations across the input patch. At the image pre-processing layer, the patches of $224\times224$ pixels are split into the non-overlapping cells of $16\times16$ pixels. Later, the linear embedding layer flattens these cells, and positional encodings are added before feeding the embeddings to the pile of transformer blocks, as illustrated in Fig.~\ref{fig:intro} (step 2). Since convolutional networks have shown their efficacy in image recognition tasks, transferring knowledge from a DCNN network can help the ViT absorb inductive biases. Therefore, we rely on popular state-of-the-art DCNNs for selecting teacher architecture. Nevertheless, we systemically discover appropriate student and teacher candidates during the experiments later to demonstrate the approach's effectiveness over TL.
\vspace{-0.5em}
\subsection{Training Student under Knowledge Distillation}
\vspace{-0.25em}
After selecting student and teacher architectures, we begin the process of training the student $\S$. The goal is to train $\S$ with the assistance of a $\T$ to improve the $\S$`s generalization performance using additional knowledge beyond the labels. Our approach is similar to Hinton \etal~\cite{hinton2015distilling} where model outputs $\s$, and $\t$ are normalized by a temperature $\temp$ parameter before using the softmax function $\sigma$. The increasing value of $\temp$ softens the impact of the fluctuations in the output probability distribution; therefore, more knowledge can be devolved with each input $\bx_{n}$. 
Instead of using softmax on $\s_{n}$, we take advantage of the log-softmax function $\sigma^*$, which stabilizes the distillation process by penalizing for incorrect class. $\sigma^*$ also adds efficiency by optimizing gradient calculations. 

The output logits for input patch $\bx_{n}$ can be written as; 
\vspace{-0.5em}
\begin{equation}\label{eq:1}
\s_{n}=\S_{\theta}(\bx_{n}) \quad \textrm{and} \quad \t_{n}=\T_{\phi}(\bx_{n})
\end{equation}

Let the log-softmax and softmax on logits, $\sigma^*(\s/\temp)$ and $\sigma(\t/\temp)$, for each element can be defined as (see Eq.~\eqref{eq:2}); 
\vspace{-0.5em}
\begin{equation} \label{eq:2}
    \sigma^*(s_{i}/\temp) = log\biggl(\frac{\exp{(s_{i}/\temp)}}{\sum_{j=1}^{c}\exp{(s_{j}/\temp)}}\biggl) \quad \textrm{and} \quad  \sigma(t_{i}/\temp) = \frac{\exp{(t_{i}/\temp)}}{\sum_{j=1}^{c}\exp{(t_{j}/\temp)}}
\end{equation}

where $c$ is the total number of classes and $\temp$ is the temperature. The class probabilities at the output of the $\S$ and $\T$ model can thus be written as;
\vspace{-0.5em}
\begin{equation} \label{eq:3}
    p_{\S}=\sigma^*(\s/\temp) = \sigma^*(\S_{\theta}(\bx)) \quad \textrm{and} \quad p_{\T}=\sigma(\t/\temp) = \sigma(\T_{\phi}(\bx))
\end{equation}

 The student loss $L_{student}$ (Eq.~\eqref{eq:student}) provides hard targets and is obtained by applying cross entropy $L_{CE}$ on ground truth $y$, and $\s$ when T is set to 1;
\vspace{-0.5em}
 \begin{equation} \label{eq:student}
    L_{student} = L_{CE}(y, \s) = -\sum_{i=1}^{c} y_{i} \cdot log(\sigma^*(s_{i}))
    \vspace{-0.5em}
\end{equation}
Distillation loss $L_{distillation}$ provides the soft targets and is computed from the $p_{\S}$ and $p_{\T}$ by applying Kullback-Leibler divergence $KL_{D}$. Since the outputs from $\S$ and $\T$ were normalized by $\temp$, we multiply the loss with $\temp^2$ to maintain their relative contribution;

\vspace{-0.5em}
\begin{equation} \label{eq:distillation}
    L_{distillation} = {T}^2 \times KL_{D} (p_{\S} \| p_{\T})= {T}^2 \cdot \sum_{i=1}^{c} p_{{\S}_{i}} log\frac{p_{{\S}_{i}}}{p_{{\T}_{i}}}
\vspace{-0.5em}
\end{equation}
The final loss function, as shown in Eq.~\eqref{eq:floss}, is a weighted average of student and distillation losses where $\alpha \in [0,1)$; 
 
\begin{equation} \label{eq:floss}
    L_{Final} = \alpha \times L_{student}  +  \beta \times L_{distillation}    \quad  \cdot :  \beta = 1-\alpha
\end{equation}
\vspace{-1em}

High entropy in soft targets offers significantly more information per training patch than hard targets~\cite{hinton2015distilling}, allowing the student ViT to train with fewer data and a higher learning rate. Therefore, using a smaller alpha can be beneficial if the $\S$ is trained from scratch. Our standalone training setup for baseline comparison can be obtained by putting $\alpha$ and $\temp$ equal to one and replacing log softmax with softmax function. 
\vspace{-0.25em}
\subsection{Prediction}
\vspace{-0.5em}
Once the final loss is minimized based on the experimental setup (defined in Sec.~\ref{sec:experiment}), we find predictions from the student $\S$ by setting $\temp$ equal to one. For an unseen test patch $\textbf{x}_*$, output can be defined as~\eqref{eq:pred};
\vspace{-0.5em}
\begin{equation} \label{eq:pred}
       \hat{y}_{s} = \argmax(\sigma(\s_{*}))=\argmax(\sigma(\S_{\theta}(\bx_{*}))) \quad   \in \ \{0,1\}
\end{equation}
\vspace{-0.5em}

\vspace{-1.5em}
\section{Experimental Setup} 

\vspace{-0.5em}\label{sec:experiment}
\vspace{-0.75em}
{\bf Implementation Details}:
The patch extraction was accomplished using the HistoLab library. Extracted patches were normalized to ImageNet~\cite{imagenet} mean and standard deviation. We augmented data at every training epoch using random geometric transformations, such as rotations, horizontal and vertical flips. ViTs were borrowed from Timm Library, and the experimental setup was built on the Pytorch. We used four variants of ViTs with different parametric depths from~\cite{dosovitskiy2020image,Deit}, where the classifier was replaced by a fully connected (FC) layer. We used four state-of-the-art DCNNs with varying parametric complexity. All DCNN backbones were initialized with ImageNet~\cite{imagenet} weights, and classifiers were replaced with three-layer FC classifiers. All classifiers were initialized with random weights.  
After hyper-parameter exploration, the final parameters were set to a batch size of 64, SGD optimizer, ReduceLROnPlateau scheduler with a learning rate of 0.001, dropout of 0.2, cross-entropy loss, and early stopping with the patience of 20 epochs on validation loss to prevent over-fitting. For KD parameters, values of $\temp \in \{2,5,10,20,40\}$ and  $\alpha\in \{0.3,0.5,0.7\}$ were explored.  
The best model weights are used to report the results. The NVIDIA GeForce A100 SXM 40GB GPU was utilized for training all models.

\vspace{0.25em}
{\bf Evaluation Metrics}: 
We evaluate the presented method using accuracy, F1-score, and Mathew Correlation Coefficient (MCC). Let TP, FN, FP, and TN describe true positive, false negative, false positive, and false negative predictions. The accuracy, termed as $(TP+TN)/(TP+FN+FP+TN)$, is the ratio of correct predictions by the model. F1 is the harmonic mean, defined as $2\cdot(\textrm{precision} \cdot \textrm{recall})/(\textrm{precision} + \textrm{recall})$ where Recall = $TP/(TP+FN)$ and Precision = $TP/(TP+FP)$. MCC is an informative measure in binary classification over imbalanced datasets and is defined as Eq.~\eqref{eq:mcc}. 
\begin{equation}\label{eq:mcc}
    \small\vspace{-0.5em}
    \begin{aligned}[c]
     MCC = \frac{TP\hspace{0.255em}\cdot\hspace{0.255em}TN - FP\hspace{0.255em}\cdot\hspace{0.255em}FN}{\sqrt{(TP+FP)\cdot(TP+FN)\cdot(TN+FP)\cdot(TN+FN)}} \quad \in [-1,1]
   \end{aligned}
\end{equation}
\vspace{-0.5em}

\vspace{-1em}
\section{Results and Discussion} \label{sec:results}

\begin{table}[htb!]
\centering\vspace{-0.3em}
\renewcommand{\arraystretch}{1.3}
\caption{\textbf{Results from Exp. 1}: Four variants of Deep Convolutional Neural Networks (DCNNs) and Vision Transformers (ViTs), with increasing parametric complexity, are trained for the air bubbles detection task. The best outcomes in every section are bolded. ViT-tiny and MobileNet architectures provide the best results on the test set.  } \vspace{-0.5em}
    \begin{tabular}{l|c| c c c |c c c}
    \hline
     \multicolumn{1}{|c}{} & \multicolumn{1}{|c}{} & \multicolumn{3}{|c|}{\textbf{Validation Set}} & \multicolumn{3}{c|}{\textbf{Test Set}} \\
    \cline{3-8}

        \multicolumn{1}{|c}{\multirow{-2}{*}{\textbf{Architecture}}} & \multicolumn{1}{|c}{\multirow{-2}{*}{\shortstack{\textbf{Param.}\\ (\#)}}} & \multicolumn{1}{|c|}{Acc.(\%)} & \multicolumn{1}{c|}{F1} & \multicolumn{1}{c|}{MCC($\Uparrow$)} & \multicolumn{1}{c|}{Acc.(\%)} & \multicolumn{1}{c|}{F1} & \multicolumn{1}{c|}{MCC($\Uparrow$)} \\
    
    \Xhline{3\arrayrulewidth}
    \multicolumn{8}{c}{\textbf{Deep Convolutional Neural Networks (DCNNs)}}\\
    \Xhline{2\arrayrulewidth}
    MobileNetv3~\cite{mobilenet} & 3.52M & 98.28 & 0.983 & 0.965 & \textbf{93.88} & \textbf{0.945} & \textbf{0.876}\\
    EfficientNet~\cite{efficientnet} & 20.89M & 96.52 & 0.966 & 0.931 & 92.54 & 0.935 & 0.851 \\
    DenseNet161~\cite{densenet} & 27.66M  & 98.12 & 0.982 & 0.962 & 91.32 & 0.925 & 0.828 \\
    VGG16~\cite{vgg16} & 136.42M & \textbf{98.34} & \textbf{0.984} & \textbf{0.966} & 92.31 & 0.932 & 0.846 \\
    
    \Xhline{3\arrayrulewidth}
    \multicolumn{8}{c}{\textbf{Vision Transformers (ViTs)}}\\
    \Xhline{2\arrayrulewidth}
    ViT-tiny~\cite{Deit} & 5.52M & \textbf{98.67} & \textbf{0.987} & \textbf{0.973} & \textbf{92.35} & \textbf{0.933} & \textbf{0.847} \\
    ViT-small~\cite{Deit} & 21.66M & 97.01 & 0.971 & 0.941 & 91.16 & 0.922 & 0.822\\
    
    ViT-large~\cite{dosovitskiy2020image} & 303.30M & 98.12 & 0.982 & 0.962 & 92.08 & 0.928 & 0.839\\
    ViT-huge~\cite{dosovitskiy2020image} & 630.76M & 95.85 & 0.962 & 0.918 & 91.43 & 0.925 & 0.829\\[0.2em]
    \hline\hline
   
    \Xhline{2\arrayrulewidth}
    \multicolumn{8}{c}{\textbf {Results from Literature (Validation Accuracy (\%))}}\\
    \Xhline{2\arrayrulewidth}
   \multicolumn{2}{l|}{DeiT-S in~\cite{raipuria2022stress} \hspace{1em}91.5-92} & \multicolumn{3}{l|}{ResNet-50 in~\cite{raipuria2022stress}  \hspace{1em}88-89} & \multicolumn{2}{l}{
  VGG16 in~\cite{Shakhawat2020}  \hspace{1em}87.33} 
    \end{tabular}
    \label{res:1}
\end{table}

\vspace{-0.5em}
 \subsection{\emph{Exp. 1}: Baseline Experiments for Architecture Decision}
\vspace{-0.5em}
In this experiment, we only apply TL to a set of architectures. We evaluate state-of-the-art DCNNs, namely MobileNetv3~\cite{mobilenet}, EfficientNet~\cite{efficientnet}, DenseNet161~\cite{densenet} and VGG16~\cite{vgg16} architectures and a family of four ViTs~\cite{dosovitskiy2020image,Deit}, with increasing architecture size. Exp 1 provides a baseline as well as helps to choose architectures for the KD setup in later experiments. Table~\ref{res:1} reports the results of the validation and test set. DCNNs largely exceed the performance of ViTs, where top-performing ViT lags the generalization performance of top-performing DCNNs by 3\% in MCC. Moreover, architectures with sizeable parameters like VGG16 and ViT-tiny and MobileNet, despite being architectures with fewer parameters, emerge as appropriate student and teacher candidates, respectively, based on the test results and outperform the results from the literature.

\vspace{-0.75em}
\subsection{\emph{Exp. 2}: How Important is Teacher`s Knowledge?}
\vspace{0.45em}

\begin{table}[hbt!]
\centering\vspace{-0.3em}
\renewcommand{\arraystretch}{1.2}
\caption{\textbf{Results from Exp. 2}: Knowledge Distillation (KD) outcome for selected teacher and student candidates from Exp.1. The values of $\alpha, T$ are fixed at 0.5 and 10, respectively. The best results in every part are marked in bold, and the second best is underlined. ViT-tiny, with two scratch and ImageNet initialization, is used for baseline comparisons. Two teachers (MobileNet and VGG16) with air bubbles knowledge are used. While MobileNet is also initialized with knowledge of other domains to evaluate the importance of teachers' knowledge. }\vspace{-0.25em}

 \begin{tabular}{l| c c c| c c c}
 \hline
 \multicolumn{1}{|c}{}  & \multicolumn{3}{|c|}{\textbf{Validation Set}} & \multicolumn{3}{c|}{\textbf{Test Set}} \\
\cline{2-7}

\multicolumn{1}{|c}{\multirow{-2}{*}{\textbf{Architecture (Initial.)}}} & \multicolumn{1}{|c}{Acc.(\%)} & \multicolumn{1}{|c}{F1} & \multicolumn{1}{|c}{MCC($\Uparrow$)} & \multicolumn{1}{|c}{Acc.(\%)} & \multicolumn{1}{|c}{F1} & \multicolumn{1}{|c|}{MCC($\Uparrow$)}\\

\Xhline{3\arrayrulewidth}
    \multicolumn{7}{l}{\textbf{Baseline (Initial.) - Standalone training}}\\
  \Xhline{2\arrayrulewidth}
  
     ViT-tiny (Scratch) & 96.13 & 0.963 & 0.922 & 91.51 & 0.925 & 0.829   \\
     ViT-tiny (ImageNet~\cite{imagenet}) & \textbf{98.67} & \textbf{0.987} & \textbf{0.973} & \textbf{92.35} & \textbf{0.933} & \textbf{0.847}\\[0.2em]
   \Xhline{3\arrayrulewidth}
    \multicolumn{7}{l}{\textbf{Teacher (Initial.) - Student [ViT-tiny (Scratch)]}}\\
    \Xhline{2\arrayrulewidth}
    
    MobileNet (Scratch) & 96.13 & 0.962 & 0.924 & 87.92 & 0.889 & 0.756 \\
    MobileNet (ImageNet~\cite{imagenet}) & 95.58 & 0.957 & 0.914 & 92.31 & 0.927 & 0.848\\
    MobileNet (Damaged~\cite{kanwal2022quantifying}) & 76.8 & 0.785 & 0.533 & 49.23 & 0.608 & -0.075 \\
    MobileNet (Air bubbles) & \textbf{98.01} & \textbf{0.981} & \textbf{0.960} & \textbf{95.25} & \textbf{0.957} & \textbf{0.904} \\
    VGG16 (Air bubbles) & \underline{97.18} & \underline{0.973} & \underline{0.944} & \underline{93.42} & \underline{0.940} & \underline{0.867} \\[0.1em]
    
     \Xhline{3\arrayrulewidth}
    \multicolumn{7}{l}{\textbf{Teacher (Initial.) - Student [ViT-tiny (ImageNet)]}}\\
    \Xhline{2\arrayrulewidth}

    MobileNet (Scratch) & \textbf{98.73} & 0.983 & 0.971 & 93.38 & 0.941 & 0.866 \\
    MobileNet (ImageNet~\cite{imagenet}) & 98.62 & \textbf{0.987} & \underline{0.972} & 93.40 & 0.942 & 0.867\\
    MobileNet (Damaged~\cite{kanwal2022quantifying}) & 50.08 & 0.211 & 0.09 & 35.51 & 0.116 & -0.294 \\
    MobileNet (Air bubbles) & 98.61 & \textbf{0.987} & \textbf{0.973} & \textbf{95.60} & \textbf{0.961} & \textbf{0.911} \\
    VGG16 (Air bubbles) & \underline{98.67} & \underline{0.986} & \underline{0.972} & \underline{94.19} & \underline{0.948} & \underline{0.882} \\[0.1em]
    \Xhline{2\arrayrulewidth}
 \end{tabular}
 \vspace{-1em}
    \label{res:2}
\end{table}
\vspace{-1em}

This experiment evaluates the impact of existing teacher knowledge in the KD process to assess the real-life analogy where good teachers make good students.  Therefore, we initialize MobileNet teachers with no knowledge (scratch), knowledge from a general domain (ImageNet), knowledge from another WSI artifact (damaged tissue~\cite{kanwal2022quantifying}), and finally, domain-relevant knowledge (air bubbles) from the previous experiment. In addition, we also select VGG16 with air bubble knowledge as a teacher to assess the effect of highly parametric DCNN in the KD process. For this experiment, the values of $\alpha, T$ are fixed at 0.5 and 10, respectively.
The student is a ViT-tiny architecture initialized with random and ImageNet weights separately. 

Table~\ref{res:2} exhibits that KD remarkably improves ViT`s classification ability. Even without ImageNet knowledge, ViT-tiny, under the KD framework, surpasses all metrics under both MobileNet and VGG16 teachers. However, the best results are obtained using the MobileNet teacher, ascertaining that hidden knowledge can be easily distilled from a simpler architecture. Interestingly, teachers with knowledge other than the relevant domain (air bubbles) produce poorly performing student. Although the student with ImageNet knowledge does not indicate gain on the validation results relative to the baseline, it achieves 3\% and 7\% improvement in F1 and MCC scores on the test set, respectively.

Overall, the test results demonstrate that the KD is promising to train generalized ViT-tiny with little data, even without pretrained weights. ViT significantly enhances its generalization against the baseline when trained in a standalone setting. Especially when the teacher is enriched with the knowledge related to the task. KD, on top of ImageNet TL, provides a marginal gain in the performance of ViT-tiny, overcoming the reliance on pretrained weights.
\vspace{-0.75em}
\subsection{\emph{Exp. 3}: Influence of KD Parameters}
\begin{figure}[htb!]
    \centering
\includegraphics[width=1\textwidth]{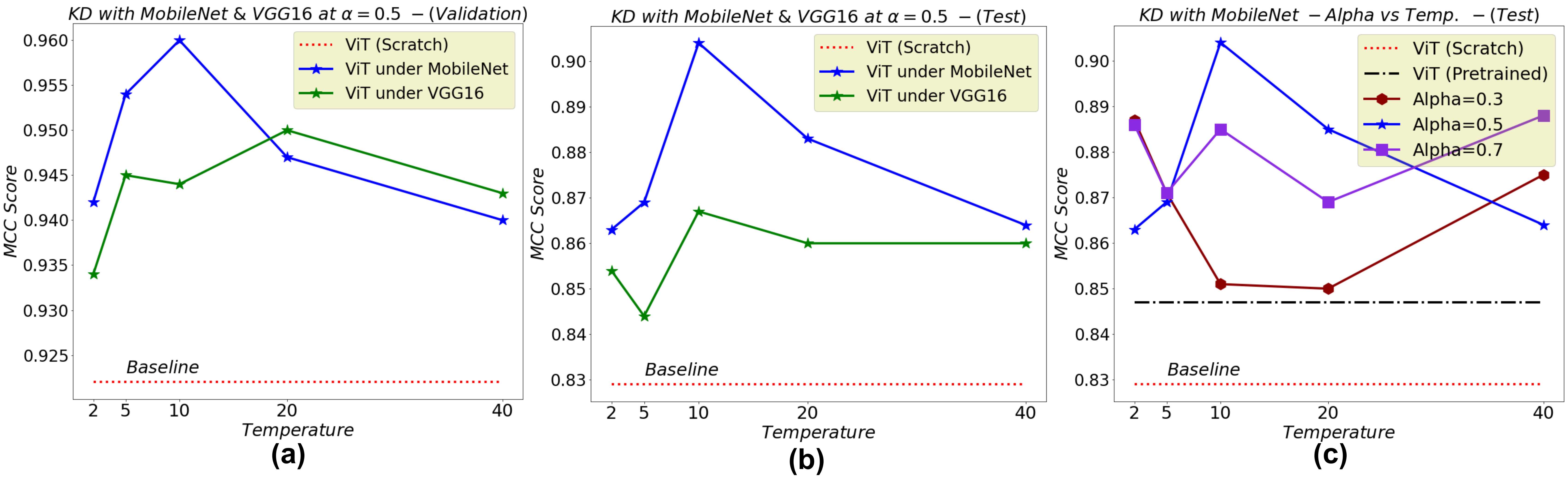}
    \vspace{-1.75em}
\caption{\textbf{Results from Exp. 3}: 
    Knowledge Distillation (KD) improves the performance of the Vision Transformer (ViT-tiny) under the supervision of both MobileNet and VGG16 teachers. (a) and (b) shows an improved performance from the baseline (standalone training from scratch), under all temperature ($T$) values, on validation and test set. (c) depicts the influence of giving higher/lower weightage to distillation loss from the teacher network (see Sec.~\ref{sec:method}). The MobileNet teacher, despite being simpler architecture, enriches ViT-tiny`s generalization capability on all chosen $\alpha$ and $T$ values.\vspace{-1em}}
    \label{fig:exp3}
\end{figure}

\vspace{-0.75em}
Since the initialization of teachers with air bubbles knowledge has been shown to improve the learning process, it would be interesting to assess the influence of DCNN teachers under the different KD parameters ($\temp$ and $\alpha$). In this experiment, we chose $\temp \in \{2,5,10,20,40\}$ and $\alpha \in \{0.3,0.5,0.7\}$ to estimate the influence of teacher`s output on ViT student, trained from scratch.  The baseline experiment corresponds to $\alpha$ and $\temp =1$ and uses sigmoid on ViT outputs.  Fig.~\ref{fig:exp3} (a) and (b) show MCC values as the effect of temperature on simple DCNN like MobileNet and complex DCNN like VGG16. Though the ViT-tiny student trained under the VGG16 teacher scores better on the validation set when $\temp$ is high, the MobileNet teacher reveals better transference of hidden knowledge on all $\temp$ values on the test set. Fig.~\ref{fig:exp3} (c) depicts the effect of $\alpha$ on ViT's generalization results. All $\alpha$ values give better results than the baseline, concluding that including distillation loss improves training compared to only student loss.  

To sum up, the teacher`s outcome strongly influences the student`s generalizability in the KD process. Most of the $\temp$ and $\alpha$ values deliver a noticeable gain over the standalone training in our case. However, \emph{intermediate} $\temp$ values and assigning \emph{equal weight} to student and distillation loss is the most advantageous. 
\vspace{-1.5em}
\section{Conclusion and Future Work} \label{sec:conclusion}
\vspace{-0.75em}
\vspace{0em}
This paper presents the Knowledge Distillation (KD) to boost the generalization performance of small Vision Transformers (ViTs) on a small histopathological dataset. The main motivation is to create a well-performing and efficient preprocessing pipeline that requires a generalized and computationally-friendly model. 
We evaluated various pretrained DCNNs and ViTs for the air bubbles artifact detection task. ViTs, trained in a standalone setting, underperform DCNNs on unseen data. Our approach exploits the KD, in the absence of pretrained weights, to enhance the performance of ViT by training under the guidance of a DCNN teacher.
Our analysis found that KD provides significant gain under most distillation settings when the teacher holds the knowledge of the same task. In conclusion, the ViT, when trained under KD, outperforms its state-of-the-art DCNN teacher and its counterpart in standalone training.  

In future work, the method can be developed and tested on larger cohorts of histological data with stain variations and to detect multiple artifacts. Moreover, artifact detection by ViT trained under the student-teacher recipe can be combined as a preprocessing step with a diagnostic or prognostic algorithm in the computational pathology system.   


\vspace{-1.25em}
\section{Acknowledgment}
\vspace{-1em}
This research is supported by the European Horizon 2020 program under Marie Skłodowska-Curie grant agreement No. 860627 (CLARIFY). The authors have no relevant financial or non-financial interests to disclose.

\vspace{-1em}
\bibliographystyle{splncs04}
\bibliography{ref}

\begin{thebibliography}{10}
\providecommand{\url}[1]{\texttt{#1}}
\providecommand{\urlprefix}{URL }
\providecommand{\doi}[1]{https://doi.org/#1}

\bibitem{bhojanapalli2021understanding}
Bhojanapalli, S., Chakrabarti, A., Glasner, D., Li, D., Unterthiner, T., Veit,
  A.: Understanding robustness of transformers for image classification. In:
  Proc. of the IEEE International Conf. on Computer Vision (ICCV). pp.
  10231--10241 (2021)

\bibitem{imagenet}
Deng, J., Dong, W., Socher, R., Li, L.J., Li, K., Fei-Fei, L.: Imagenet: A
  large-scale hierarchical image database. In: 2009 IEEE ICCV. pp. 248--255.
  Ieee (2009)

\bibitem{dosovitskiy2020image}
Dosovitskiy, A., Beyer, L., et~al.: An image is worth 16x16 words: Transformers
  for image recognition at scale. arXiv preprint arXiv:2010.11929  (2020)

\bibitem{fuster2022invasive}
Fuster, S., Khoraminia, F., et~al.: Invasive cancerous area detection in
  non-muscle invasive bladder cancer whole slide images. In: IEEE 14th Image,
  Video, and Multidimensional Signal Processing Workshop (IVMSP). pp.~1--5.
  IEEE (2022)

\bibitem{geirhos2018imagenet}
Geirhos, R., Rubisch, P., Michaelis, C., Bethge, M., Wichmann, F.A., Brendel,
  W.: Imagenet-trained cnns are biased towards texture; increasing shape bias
  improves accuracy and robustness. arXiv preprint arXiv:1811.12231  (2018)

\bibitem{golatkar2018classification}
Golatkar, A., Anand, D., et~al.: Classification of breast cancer histology
  using deep learning. In: International conf. image analysis and recognition.
  pp. 837--844. Springer (2018)

\bibitem{guan2021mri}
Guan, H., Wang, C., Tao, D.: Mri-based alzheimer’s disease prediction via
  distilling the knowledge in multi-modal data. NeuroImage  \textbf{244},
  118586 (2021)

\bibitem{he2021self}
He, J., Li, C., Ye, J., Qiao, Y., Gu, L.: Self-speculation of clinical features
  based on knowledge distillation for accurate ocular disease classification.
  Biomedical Signal Processing and Control  \textbf{67},  102491 (2021)

\bibitem{resnet}
He, K., Zhang, X., Ren, S., Sun, J.: Deep residual learning for image
  recognition. In: Proceedings of the IEEE CVPR. pp. 770--778 (2016)

\bibitem{hinton2015distilling}
Hinton, G., Vinyals, O., Dean, J., et~al.: Distilling the knowledge in a neural
  network. arXiv preprint arXiv:1503.02531  \textbf{2}(7) (2015)

\bibitem{Hossain2018}
Hossain, M.S., Nakamura, T., Kimura, F., Yagi, Y., Yamaguchi, M.: Practical
  image quality evaluation for whole slide imaging scanner. In: Biomedical
  Imaging and Sensing Conference. vol. 10711, pp. 203--206. SPIE (2018)

\bibitem{mobilenet}
Howard, A., Sandler, M., Chu, G., Chen, L.C., Chen, B., Tan, M., Wang, W., Zhu,
  Y., Pang, R., Vasudevan, V., et~al.: Searching for mobilenetv3. In:
  Proceedings of the IEEE International Conference on Computer Vision. pp.
  1314--1324 (2019)

\bibitem{densenet}
Huang, G., Liu, Z., Van Der~Maaten, L., Weinberger, K.Q.: Densely connected
  convolutional networks. In: Proceedings of the IEEE conference on computer
  vision and pattern recognition. pp. 4700--4708 (2017)

\bibitem{kanwal2023detection}
Kanwal, N., Amundsen, R., Hardardottir, H., Janssen, E.A., Engan, K.: Detection
  and localization of melanoma skin cancer in histopathological whole slide
  images. arXiv preprint arXiv:2302.03014  (2023)

\bibitem{kanwal2022quantifying}
Kanwal, N., Fuster, S., et~al.: Quantifying the effect of color processing on
  blood and damaged tissue detection in whole slide images. In: IEEE 14th
  Image, Video, and Multidimensional Signal Processing Workshop (IVMSP).
  pp.~1--5. IEEE (2022)

\bibitem{kanwal2022devil}
Kanwal, N., P{\'e}rez-Bueno, F., Schmidt, A., Engan, K., Molina, R.: The devil
  is in the details: Whole slide image acquisition and processing for artifacts
  detection, color variation, and data augmentation. IEEE Access  \textbf{10},
  58821--58844 (2022)

\bibitem{kanwal2022attention}
Kanwal, N., Rizzo, G.: Attention-based clinical note summarization. In:
  Proceedings of the 37th ACM Symposium on Applied Computing. pp. 813--820
  (2022)

\bibitem{lingmei2021noninvasive}
Lingmei, A., et~al.: Noninvasive grading of glioma by knowledge distillation
  base lightweight convolutional neural network. In: IEEE 2021 AEMCSE. pp.
  1109--1112

\bibitem{meng2021knowledge}
Meng, H., Lin, Z.e.a.: Knowledge distillation in medical data mining: A survey.
  In: 5th International Conf. on Crowd Science and Engineering. pp. 175--182
  (2021)

\bibitem{naseer2021intriguing}
Naseer, M., Ranasinghe, K., Khan, S., Hayat, M., Shahbaz~Khan, F., Yang, M.H.:
  Intriguing properties of vision transformers. NeurIPS  \textbf{34},
  23296--23308 (2021)

\bibitem{noorbakhsh2020deep}
Noorbakhsh, J., Farahmand, S., et~al.: Deep learning-based
  cross-classifications reveal conserved spatial behaviors within tumor
  histological images. Nature communications  \textbf{11}(1),  1--14 (2020)

\bibitem{raipuria2022stress}
Raipuria, G., Singhal, N.: Stress testing vision transformers using common
  histopathological artifacts. In: Medical Imaging with Deep Learning (2022)

\bibitem{salehi2021multiresolution}
Salehi, M., Sadjadi, N., Baselizadeh, S., Rohban, M.H., Rabiee, H.R.:
  Multiresolution knowledge distillation for anomaly detection. In: Proceedings
  of the IEEE conference on computer vision and pattern recognition. pp.
  14902--14912 (2021)

\bibitem{Shakhawat2020}
Shakhawat, H.M., Nakamura, T., Kimura, F., Yagi, Y., Yamaguchi, M.: Automatic
  quality evaluation of whole slide images for the practical use of whole slide
  imaging scanner. ITE Trans. on Media Technology and Applications
  \textbf{8}(4),  252--268 (2020)

\bibitem{vgg16}
Simonyan, K., Zisserman, A.: Very deep convolutional networks for large-scale
  image recognition. arXiv preprint arXiv:1409.1556  (2014)

\bibitem{efficientnet}
Tan, M., Le, Q.: Efficientnet: Rethinking model scaling for convolutional
  neural networks. In: International conf. on machine learning. pp. 6105--6114.
  PMLR (2019)

\bibitem{taqi2018review}
Taqi, S.A., Sami, S.A., Sami, L.B., Zaki, S.A.: A review of artifacts in
  histopathology. Journal of oral and maxillofacial pathology: JOMFP
  \textbf{22}(2), ~279 (2018)

\bibitem{tomasetti2022multi}
Tomasetti, L., Khanmohammadi, M., Engan, K., H{\o}llesli, L.J., Kurz, K.D.:
  Multi-input segmentation of damaged brain in acute ischemic stroke patients
  using slow fusion with skip connection. arXiv preprint arXiv:2203.10039
  (2022)

\bibitem{Deit}
Touvron, H., et~al.: Training data-efficient image transformers \& distillation
  through attention. In: Int. Conf. on Machine Learning. pp. 10347--10357
  (2021)

\bibitem{vaswani2017attention}
Vaswani, A., Shazeer, N., et~al: Attention is all you need. Advances in neural
  information processing systems  (2017)

\end{thebibliography}

\end{document}